\documentclass[runningheads, orivec]{llncs}
\usepackage{graphicx}
\usepackage{amsmath}
\usepackage{amssymb}
\usepackage{float}
\usepackage[pagebackref=true,breaklinks=true,colorlinks,bookmarks=false]{hyperref}

\begin{document}
\title{CALVIS: chest, waist and pelvis circumference from 3D human body meshes	
as ground truth for deep learning}
\titlerunning{Chest, waist and pelvis circumference from 3D human body meshes}
\author{Yansel Gonzalez-Tejeda\orcidID{0000-0003-1002-3815} \and
	Helmut Mayer\orcidID{0000-0002-2428-0962}}
\authorrunning{Gonzalez-Tejeda and Mayer}
\institute{Paris Lodron University of 
	Salzburg, Kapitelgasse 4-6, 5020 Salzburg, Austria
	\url{https://www.uni-salzburg.at}}
\maketitle              % typeset the header of the contribution
\begin{abstract}
In this paper we present CALVIS, a method to calculate \textbf{C}hest, 
w\textbf{A}ist and pe\textbf{LVIS} circumference from 3D human body meshes. Our 
motivation is to use 
this data as ground truth for training convolutional neural networks 
(CNN). Previous work had used the large scale CAESAR dataset or determined 
these anthropometrical measurements $\textit{manually}$ from a person 
or human 3D body meshes. 
Unfortunately, acquiring these data is a cost and 
time consuming endeavor. In contrast, our method can be used on 
3D meshes automatically. We synthesize eight human body meshes and 
apply 
CALVIS to calculate chest, waist and pelvis circumference. We evaluate the 
results qualitatively and observe that the measurements can indeed be used 
to estimate the shape of a person. We then asses the 
plausibility of our approach by generating
ground truth with CALVIS to train a small CNN. After
having trained the network with 
our data, we achieve competitive validation error.
Furthermore, we make the implementation of CALVIS publicly available to 
advance the field.

\keywords{chest, waist and pelvis circumference \and 3D human body mesh \and 
deep learning}
\end{abstract}
\section{Introduction}\label{sec:intro}
\textbf{Motivation} 
Predicting 3D human body measurements from images is crucial in several 
scenarios 
like virtual try-on, animating, ergonomics, computational forensics and 
even health and mortality estimation. Researches had 
named these
measurements body intuitive controls \cite{Allen.2003}, biometric 
measurements \cite{Sigal.2008}, body dimensions 
\cite{DBLP:conf/bmvc/ChenRC11}, (European standard EN 13402), semantical 
parameters \cite{Yang.2014}, 
traditional anthropometric 
measurements 
\cite{Wuhrer2011} or only ``shape" as in human shape estimation 
\cite{Guan.2013,Bogo:ECCV:2016,Loper.2015,Dibra.2016a,Pishchulin.2017}. In 
contrast, we assume a more 
anthropometric approach 
motivated by comprehensive compendiums like Panero and Zelnik, 1979 
\cite{panero1979human}.
Throughout this paper the term human body 
dimensions 
will be used to refer to the above measurements and we will focus on three of 
these dimensions, namely chest, waist and 
pelvis circumference.

The problem of estimating chest, waist and 
pelvis circumference having only an image of a person is a challenging problem 
and involve recognizing the corresponding location of these dimensions in the 
body and estimating their circumference. Additionally, the setting is an  
under-constrained  (or inverse) problem. Information 
gets lost when a camera is 
used to capture the human body in 3D space to 'render' a 2D image. 

To tackle this problem a supervised learning approach can be used. This approach
demands large amount of human body anthropometric measurements and is certainly 
one of the biggest challenges in the field.  Currently there 
is only one large-scale dataset, the Civilian American and 
European Surface Anthropometry Resource (CAESAR) \cite{robinette1999caesar} 
with 3D human body scans and their corresponding anthropometric measurements. 
This survey was extraordinarily large and resource intensive: around nine 
thousand people from Europe and the USA where scanned, it was conducted over 
more than five years and costed more than six million dollars.

In the past decade a noticeable number of researchers have employed this 
dataset to investigate human shape estimation. Because the measurement 
acquiring process is resource intensive and requires large amount of human and 
material resources, this type of studies are rare and the data they offer is 
expensive. Therefore, it is important to explore alternative methods where 
human body measurements derived from real data can be obtained for 
investigation.

3D human body generative models offers such an alternative. In the line of 
previous work, we start by synthesizing 3D human body meshes using the Skinned 
Multi-Person Linear (SMPL) \cite{Loper.2015} generative 
model. Unlike other works, once we have 
the 3D meshes we compute chest, waist and pelvis circumference. The next 
step after obtaining the 
measurements is to use a camera model to render
images. Finally, in possession of this ground truth we can 
input these images to the learning algorithm and train with the human body dimensions as supervision signal.

\textbf{Contributions} In summary, our contributions are $1)$ CALVIS: a method 
capable of automatically outputting chest, waist and pelvis circumference from 
3D body meshes. $2)$ A prototype deep learning framework (CNN) with synthetic 
images of 3D human body meshes as input and the human body dimensions obtained 
with CALVIS as supervision signal.

\section{Related Work} \label{sec:related_work}

%-------------------------------------------------------------------------
\subsection{Human Dimensions Estimation from Images.}
Using metrology techniques \cite{ams.2008.BenAbdelkaderY08} and mixture of 
experts model \cite{Sigal.2008} human body dimensions like height, arm span and 
even weight had been estimated from images. Chest and waist size had been 
considered as well \cite{Guan.2013}. These investigations used human dimensions 
for validation (ground truth) that were directly recovered from individuals. In 
contrast, we calculate them  
consistently from 3D meshes.
Another significant research direction studies shape as (non necessarily meaningful) parameters of some model. For example \cite{Bogo:ECCV:2016} claim describing the first method 
to 
estimate pose and shape from a single image, while \cite{DBLP:Lassner2017} estimates shape from 91 keypoints using decision 
forests. Shape is understood as the 
parameters of the SMPL model but not human dimensions.

In general, previous work has concentrated on human body shapes not deviating 
much from the mean. In contrast, we move beyond by exploring 3D meshes 
reflecting human figures characteristics such as bulky, slim, small and 
tall subjects.

\subsection{Human Body Data Synthesis.}
The Shape Completion and Animation of People (SCAPE) model\cite{Anguelov.2005} 
opened wide possibilities in the field of 
human shape estimation. It 
provided the scientific community with a method to build realistic 3D human 
body meshes of different shapes in different poses. In order to synthesize a 
human body mesh a template body must be deformed. The model exhibits, however, an important limitation, namely the last 
step on the pipeline is an expensive optimization problem. Other researchers had used variation of the SCAPE model (S-SCAPE \cite{Pishchulin.2017}) to synthesize human bodies but focused on people detection.

After some attempts on improving the human models quality, for example, to make 
more 
easily capturing the correlation between body shape and pose 
\cite{HaslerSSRS09}, or to better understand body dimensions (Semantic 
Parametric Reshaping of Human Body Models - SPRING model 
\cite{Yang.2014}), Loper et al., 2015 \cite{Loper.2015} developed the SMPL generative 
human body model. In this work we synthesize 3D human meshes using this 
model. We will briefly describe the model in subsection \ref{subsec:smpl_model}.

Our approach to synthesize images from 3D meshes has been influenced 
by the recent publication of the large scale dataset SURREAL 
\cite{varol17_surreal}. This work uses the SMPL model 
to generate images 
from humans with random backgrounds. However, no human body dimensions are 
computed or estimated.

More recently, a human model with added hands and face had been presented \cite{Joo.2018}. We do not use this more complex model 
because the human dimensions we estimate do not require such level 
of detail.

\subsection{Human Body Dimensions from 3D Models.}
Although extensive research has been carried out on human shape estimation, 
methods to consistently define shape based on 3D mesh 
anthropometric measurements have been little explored. Only a handful of researchers 
have 
reported calculating 1D body dimensions from 3D triangular meshes to use them as ground truth for training and validation in a variety of inference processes.
 
Early research performed feature analysis on what they 
call body intuitive controls, e.g. height, weight, age and gender \cite{Allen.2003}  but they do not 
calculate them. Instead they use the CAESAR demographic data. Recording human dimensions beyond height like body fat and the more abstract ``muscles" are described by \cite{HaslerSSRS09}.

Pertinent to our investigation is also the inverse problem: generating 3D human 
shapes from traditional anthropometric measurements \cite{Wuhrer.2013}. Like this work we use a set of 
anthropometric measurements that we call dimensions, unlike them we calculate 
1D measurements from 3D human bodies to use them as ground truth for later 
inference.

Strongly related to our work are methods that calculate waist and 
chest circumference by slicing the mesh on a fixed plane and compute the convex hull of the contour \cite{Guan.2013} or path length from identified (marked) 
vertices on the 3D mesh \cite{Boisvert.2013,Dibra.2016a,Dibra.2016b}. However, is not clear how they define the human body dimensions.

In contrast, we do not 
calculate the dimensions from fix vertices on the template 
mesh. Instead, we 
take a more anthropometric approach and use domain knowledge to calculate these 
measurements.

\subsection{Human Shape Estimation with Artificial Neural Networks}
A huge amount of research has been conducted in recent years to address the 
problem of 3D/2D human shape estimation using CNNs. A state-of-the-art method 
is \cite{kanazawaHMR18} where they estimate human 
shape and pose. While these CNN's output are human body models parameters 
(i.e., $\beta$s in \cite{Dibra.2016a} and \cite{varol18_bodynet}) our network 
is 
capable to output human dimensions directly.

%------------------------------------------------------------------------
\section{Chest, Waist and Pelvis Measurements}\label{sec:approach}
While human body 
meshes are traditionally acquired by specialized 3D scanners, we use a 
generative model to synthesize them, which we briefly review in 
subsection \ref{subsec:smpl_model}. To calculate the actual human body 
dimensions, we first need to formalize the 
notion of chest, waist and pelvis of a 3D human body mesh. Our strategy consist 
of segmentating the mesh in five regions, 
three of them of interest, which we 
discuss in subsection \ref{subsec:three_regions}. Finally, within these regions 
we identify the intended dimensions employing a human body mesh signature 
(HBMS), 
defined in subsection \ref{subsec:hbm_signature}.

\subsection{SMPL Generative Model of the Human Body}\label{subsec:smpl_model}
In this work we synthesize 3D human meshes using the Skinned Multi-Person 
Linear (SMPL) model \cite{Loper.2015}. The generative nature of our approach 
establishes this 
model as starting point (and not the 3D mesh). Nevertheless, our method is 
flexible enough to begin the pipeline with a 3D mesh. In that case, an SMPL 
model can be fitted to the volumetric data, using the method described by 
Varol et al., 2018 \cite{varol18_bodynet} for example.

The SMPL model is at its top level a \textbf{skinned articulated 
model}, i.e., 
consists of a 
surface mesh $\mathcal{M}$ that mimics the skin and a corresponding skeleton 
$\mathbf{J}$. The mesh $\mathcal{M}$, which is a boundary 
representation stores 
both the body geometry (vertex position) and topology (how the vertices are 
connected). The skeleton $\mathbf{J}$ is defined by its joints location in 3D 
space $j_i \in \mathbb{R}^3$  and its kinematic tree. Two 
connected 
joints define a `bone'. Moreover, a child bone is rotated relative to 
its 
connected parent. The pose $\vec{\theta}$ is the specification of every bone 
rotation plus an orientation for the root bone.

The SMPL model is also a \textbf{deformable model} 
\cite{Terzopoulos.1987}. In order to 
synthesize a 
new human mesh one has to deform the provided template mesh by 
setting shape parameters $\vec{\beta}$. Pose parameters $\vec{\theta}$ are used 
for animation.

More specifically, the model is defined by a template mesh (mean template 
shape) $\mathbf{\bar{T}}$ represented by a vector of $N = 6890$ concatenated 
vertices in the zero pose, $\vec{\theta}^*$. The skeleton has $K = 23$ joints.

As an \textbf{statistical model}, SMPL provides learned parameters
\begin{equation} \label{eq:smpl_params}
\Phi = \{\mathbf{\bar{T}}, \mathcal{W}, \mathcal{S}, \mathcal{J}, 
\mathcal{P}\}.
\end{equation}

As mentioned above $\mathbf{\bar{T}} \in \mathbb{R}^{3N}$  is the mean template 
shape. The set of blend weights $\mathcal{W} \in \mathbb{R}^{N \times K}$ 
represents 
how much the rotation of 
skeleton bones affects
the vertices. In addition, the matrices $\mathcal{S}$ and $\mathcal{P}$ define 
respectively linear functions $B_s(\vec{\beta}; \mathcal{S}): 
\mathbb{R}^{|\vec{\beta}|} \to 
\mathbb{R}^{3N}$ and $B_p(\vec{\theta}; \mathcal{P}): 
\mathbb{R}^{|\vec{\theta}|} \to 
\mathbb{R}^{3N}$ that are used to deform $\mathbf{\bar{T}}$; and the function 
$\mathcal{J}: \mathbb{R}^{|\vec{\beta}|} \to \mathbb{R}^{3K}$ predicts skeleton 
rest joint locations from vertices in the rest 
pose. A new mesh $\mathcal{M}_{new}$ can be then generated using the SMPL model 
$M$

\begin{equation}\label{eq:gen_mesh}
\mathcal{M}_{new} = M(\vec{\beta}, \vec{\theta}, \Phi).
\end{equation}

Since we held fix parameters $\Phi$ during the synthesis and either we change 
$\vec{\theta}$ because our approach focuses on the zero pose $\vec{\theta}^*$, 
we manufacture a new mesh

\begin{equation}\label{eq:gen_mesh_only_shape}
\mathcal{M}_{new} = M(\vec{\beta}).
\end{equation}

\begin{equation}\label{eq:gen_new_mesh}
\mathcal{M}_{new} = \mathbf{\bar{T}} + B_s(\vec{\beta}; \mathcal{S}).
\end{equation}

As a result we obtain a mesh that realistically represents a human body. Next, 
we focus on establishing reasonable chest, waist and pelvis regions on this 
mesh.

\subsection{Chest, Waist and Pelvic Regions 
Segmentation}\label{subsec:three_regions}

Let us consider a human body mesh $\mathcal{M}$. Our method requires that 
$\mathcal{M}$ is standing with arms raised 
parallel to the 
ground at shoulder height at a $90^\circ$ angle (SMPL zero pose
$\vec{\theta}^*$). Additionally, we assume that the mesh 
has LSA orientation, e.g., x, y and z axis are positively directed from 
right-to-left, inferior-to-superior and posterior-to-anterior, respectively. If 
the mesh has another orientation we can always apply rigid transformations to 
LSA-align it.

\begin{figure}[t]
	\begin{center}
		\includegraphics[width=\linewidth]{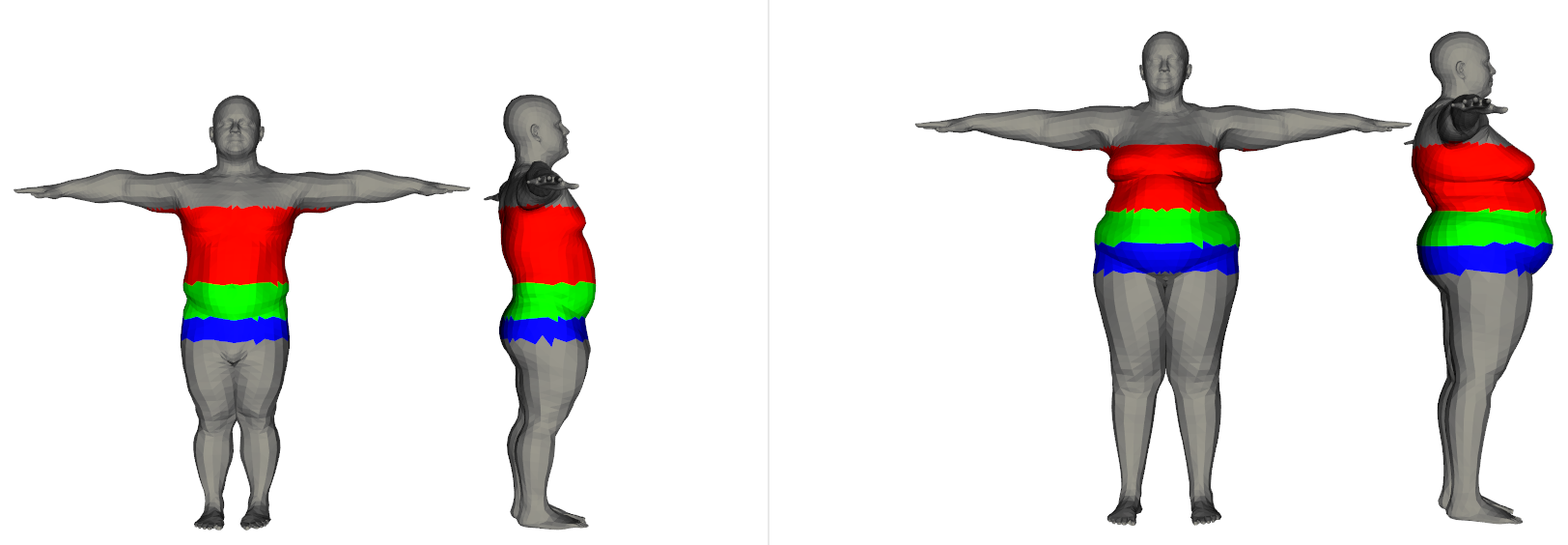}
	\end{center}
	\caption{Mesh segmentation in chest (red), waist (green) and pelvic (blue) 
	regions. We show male (left) and female (right) subjects in frontal and 
	lateral view. Note how the regions are correctly segmented for different 
	gender and body shapes.}
	\label{fig:mesh_segmented}
\end{figure}

We observe that the chest circumference is usually measure below the armpits, 
also known as axilla, but above the natural waist. Similarly, the pelvis 
circumference is measured often around the hips. This observation suggests that 
we should consider segmenting the mesh in meaningful regions. Moreover, we can 
use the skeleton when determining the region boundaries. For 
example, there is consensus regarding the axilla 
definition as the area of the human body underneath the shoulder joint 
(\cite{MeSH.axilla}, \cite{FMA.axilla}, \cite{TA.axilla}). Therefore, we can 
use the shoulder joint as a hint to locate the axilla and establish a upper 
bound for the chest region. More generally, we can use the 
skeleton joints defined in $\mathbf{J}$ to identify the regions boundaries.

Let the axilla location be the point $p_a = (p^x_a, p^y_a, p^z_a)$ and 
$j_{nw}$, $j_{pl}$, 
$j_{rh} \in \mathbf{J}$ be the natural waist, pelvis and right hip joints 
(joints with identifier $\mathtt{Spine1}$, $\mathtt{Pelvis}$, $\mathtt{R\_Hip}$ 
in 
$\mathbf{J}$), respectively.

Recall that mesh $\mathcal{M}$ has $N$ vertices. Let the set of vertices be 
$\mathcal{V}$ and $v^y_i$ the y-coordinate of vertex $v_i$, we can 
segment the 
mesh in chest, waist and pelvic regions 
$\mathcal{CR}, \mathcal{WR}, \mathcal{PR}$, respectively

\begin{align}\label{eq:cr}
\mathcal{CR} = \big(v^y_i \, | \, p^y_a > v^y_i \geq j^y_{nw} \big)
\end{align}

\begin{align}\label{eq:wr}
\mathcal{WR} = \big(v^y_i \, | \, j^y_{nw} > v^y_i \geq j^y_{pl}\big)
\end{align}

\begin{align}\label{eq:pr}
\mathcal{PR} = \big(v^y_i \, | \, j^y_{pl} > v^y_i \geq j^y_{rh}\big)
\end{align}

Figure \ref{fig:mesh_segmented} shows the result of applying these equations on 
two meshes. Note how the regions are correctly segmented for different gender 
and body shapes. For example, the male subject on the left is smaller than the 
female subject on the right. Additionally, the subjects have very different 
muscular and fat build.

\subsubsection{Axilla recognition}\label{subsec:armpit_recog}
As mentioned above, we would like to measure the chest circumference below the 
armpit, also known as axilla. This raises the question how we can recognize 
the axilla. What we want is a point $p_a$ in the mesh at the proper location 
under 
the arms. One point suffices because we require only a reference $p^y_a$ along 
the 
y-axis to define the chest region according to equation \ref{eq:cr}. Therefore, 
we focus on the right shoulder joint $j_{rs} \in \mathbf{J}$. Since the axilla 
is related to 
the shoulder joint, we can cast a ray $\boldsymbol{\iota}$ from it in direction 
to the middle left 
inferior edge of the bounding box outside the mesh at point

\begin{figure}[t]
	\begin{center}
		\includegraphics[width=\linewidth]{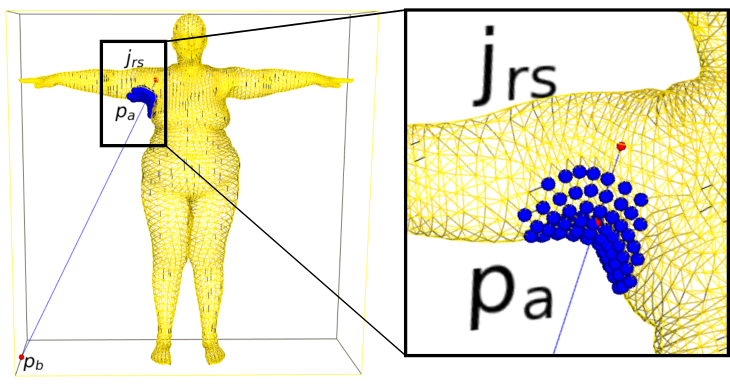}
	\end{center}
	\caption{Axilla recognition based on the skeleton joint. We cast a ray from 
		the right shoulder joint $j_{rs}$ (red point inside the mesh) in direction to the middle left 
		inferior edge of the bounding box (left). The ray intersects the mesh 
		at point $p_c$ (red) that we consider the axilla center. We plot the 80 
		nearest vertices to $p_c$ (right). Among these vertices point $p_a$ is the vertex 
		with smallest y coordinate. As it can be observed, the axilla is 
		properly recognized.}
	\label{fig:axilla-recognition}
\end{figure}

\begin{align}
p_b = \bigg(v^x_{min}, v^y_{min}, v^z_{min} + 
\frac{|v^z_{max} - v^z_{min}|}{2}\bigg).
\end{align}

Here $v^x_{min}, v^y_{min}, v^z_{min}$ are vertex minimum $x$,$y$ and $z$ 
coordinates, respectively; analog is $v^z_{max}$ the maximum vertex $z$ 
coordinate. The ray 
$\iota$ is then determined by points $j_{rs}$ (inside the mesh) and $p_b$ 
(outside the mesh). Therefore, the ray $\iota$ intersects the mesh at one point 
$p_c$ (Figure \ref{fig:axilla-recognition}), which we consider the axilla 
center. Furthermore, we can consider an strategy to increase the method 
robustness against the large variability of the human body. Since this approach 
relies on one point to establish the chest upper bound, we identify the 80 
nearest neighbors in the set of vertices $\mathcal{V}$ to $p_c$ and define 
$p_a$ to be the vertex with the smallest y coordinate of the neighbors.  

Once we have a segmented mesh, we search the dimensions within 
these regions. Next subsection discuss how to calculate geodesic on the mesh to 
establish a search strategy.

\subsection{Human Body Mesh Signature}\label{subsec:hbm_signature}
Intuitively, we would like to measure the chest circumference below the arms 
at the \textit{widest} part of the torso and the waist circumference at the 
\textit{narrowest} part beneath the chest but above the hips. This intuition is 
compliant 
with prior body dimensions standardized definition, for example, the European 
standard EN 13402-1 \cite{en13402-1}. Similarly, the pelvis 
circumference is measured often around the \textit{widest} part of hips and 
buttocks. The general idea is to cast the chest, waist and pelvis circumference 
estimation problem as a constrained optimization problem. If we are able to 
establish a 
function that measures meaningful geodesics on the mesh, we can search for its 
maximum, 
e.g., chest in the chest region.

\begin{figure}[H]
	\begin{center}
		\includegraphics[width=0.5\linewidth]{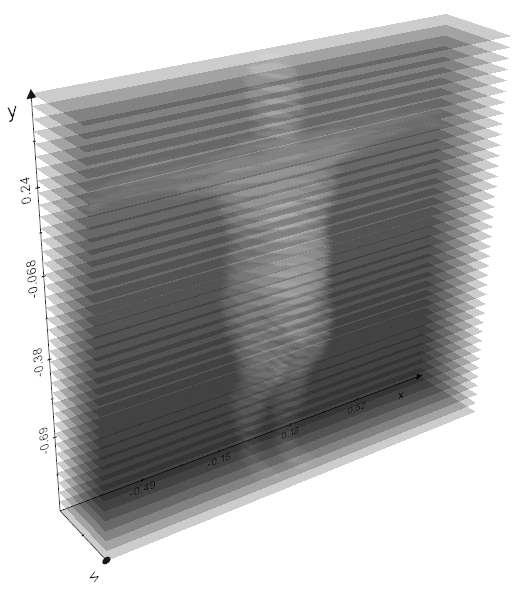}
	\end{center}
	\caption{3D mesh iterative slicing. We 
		can use a plane $\boldsymbol{\pi_j}$ parallel (with normal 
		$\vec{n}_{||}$) to the 
		floor to intersect the mesh at point $q_j = (0, q^y_j, 0), \,  q^y_j 
		\in 
		\mathbb{R}$ along the y-axis. By varying point $q_j$ (we show here 40), 
		we can obtain 
		slices.}
	\label{fig:subjects_planes_xrays}
\end{figure}

We 
can formalize this intuition by considering the cross-sectional length of the 
2D intersection curves along the y-axis. Moreover, we 
can 
use a plane $\boldsymbol{\pi_j}$ parallel (with normal $\vec{n}_{||}$) to the 
floor to intersect the mesh at point $q_j = (0, q^y_j, 0), \,  q^y_j \in 
\mathbb{R}$ along the y-axis. Since $\mathcal{M}$ is triangulated, the boundary 
of this 
intersection is a polygonal curve consisting in segments with length $s_i$. 
Therefore, we 
can 
determine the intersection boundary length as
\begin{equation}\label{eq:boundary_length}
\mathcal{BL}(\mathcal{M}, \vec{n}_{||}, q) = \sum_{i = 
	1}^{i = N}s_i
\end{equation}

Starting 
from the 
top $q_t$ of the bounding box we slice iteratively mesh $\mathcal{M}$ with 
plane 
$\boldsymbol{\pi_j}$ every $m$-meters along the y-axis until we reach the 
bounding box bottom 
$q_b$. Figure \ref{fig:subjects_planes_xrays} shows 40 of these 
slices. Next, we assemble the mesh slicing vector $\vec{\mathcal{L}}$. The 
elements of 
this vector are slice points

\begin{align}\label{eq:mesh_slice_vector}
\vec{\mathcal{L}}(\mathcal{M}, \vec{n}_{||}, m) = \left[ q_t, q_t-m, q_t-2 
m,\cdots, 
q_b \right]
\end{align}

Finally, we apply equation \ref{eq:boundary_length} to every slice defined in
$\mathcal{L}$ in equation \ref{eq:mesh_slice_vector}. In other words, for every 
slice at point 
$q_j$, we compute intersection boundary length $\mathcal{BL}(q_j)$. Here we 
drop from the notation mesh 
$\mathcal{M}$ and plane normal $\vec{n}_{||}$ because they remain constant. We 
then can 
define the \textbf{human body mesh signature} $\mathcal{MS}: 
\vec{\mathcal{L}} \to \mathbb{R}$ that maps every slice at point $q_j$ to the 
corresponding boundary length $\mathcal{BL}(q_j)$.

\begin{align}\label{eq:mesh_signature_short}
\mathcal{MS}(\vec{\mathcal{L}}) = [\mathcal{BL}(\mathcal{L}_1), 
\mathcal{BL}(\mathcal{L}_2), \cdots, 
\mathcal{BL}(\mathcal{L}_{|\vec{\mathcal{L}}|})]
\end{align}

\begin{align}\label{eq:mesh_signature}
\mathcal{MS}(\mathcal{M}, \vec{n}_{||}, m) = 
[\mathcal{BL}(q_t), \mathcal{BL}(q_t-m), \cdots, \mathcal{BL}(q_b)]
\end{align}

\begin{figure}[t]
	\begin{center}
		\includegraphics[width=\linewidth]{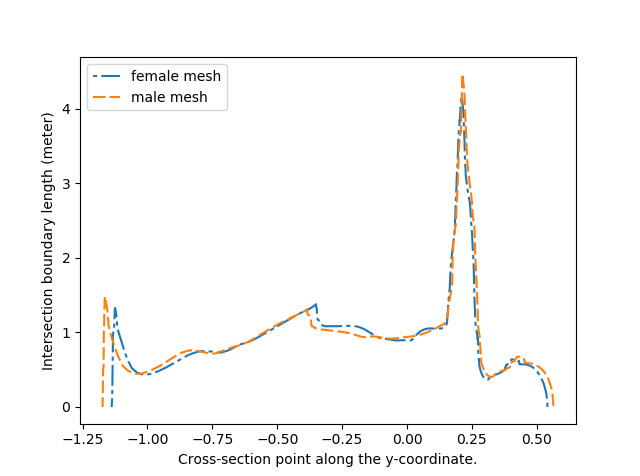}
	\end{center}
	\caption{Human Body Mesh Signature for male and female meshes. The 
		function resembles a rotated silhouette of the human body and exhibits 
		several \textit{extrema}.}
	\label{fig:hbm_signature}
\end{figure}

Figure \ref{fig:hbm_signature} shows the $\mathcal{MS}$ of two meshes (male 
and female) for 
$m=0,001$ and plane parallel to the floor (with normal 
$\vec{n}_{||} = (0, 
1,0)$) using the library trimesh \cite{trimesh}. Note that the function as 
defined by equation \ref{eq:mesh_signature} is bounded and not continuous. It 
resembles a rotated silhouette of the human body and exhibits several 
\textit{extrema}.

In general, we expect these \textit{extrema} to be adequate features to 
calculate the human dimensions. More specifically, we assume that:
\begin{enumerate}
	\item The chest circumference $cc$ is the local maximum within the chest 
	region.
	\item The waist circumference $wc$ (not to be confused with the natural 
	waist line based on the joint, see subsection \ref{subsec:three_regions}) 
	is the minimum within the waist region.	
	\item The pelvis circumference $pc$ is the maximum within the pelvis region.
\end{enumerate}

\begin{align}\label{eq:cc}
cc = \underset{\mathcal{BL}(q_j)}{\text{arg max}}\ \mathcal{MS} \quad 
\text{subject 
to} \quad p^y_a > q_j \geq j^y_{nw}
\end{align}

\begin{align}\label{eq:wc}
wc = \underset{\mathcal{BL}(q_j)}{\text{arg min}}\ \mathcal{MS} \quad 
\text{subject 
	to} \quad j^y_{nw} > v^y_i \geq j^y_{pl}
\end{align}

\begin{align}\label{eq:pc}
pc = \underset{\mathcal{BL}(q_j)}{\text{arg max}}\ \mathcal{MS} \quad 
\text{subject 
	to} \quad j^y_{pl} > v^y_i \geq j^y_{rh}
\end{align}

%------------------------------------------------------------------------
\section{Experiments and Results}

We conduct two experiments. In the first experiment we synthesize eight (four 
female and four male) human body meshes using shape parameters provided by 
SURREAL \cite{varol17_surreal}.
The meshes reflect human figures characteristics such as bulky, slim, small and 
tall subjects. Then we apply CALVIS to calculate chest, waist and pelvis 
circumference. Since we do not have ground truth, we evaluate the results 
qualitatively.
The second 
experiment serves to asses the plausibility of our approach to use the 
synthetic data for deep learning.

\subsection{Qualitative evaluation}\label{subsec:qualitative_eval}
A qualitative evaluation is in this case pertinent. There is currently no benchmark to measure how accurate the chest, waits and pelvis circumference can be. The main issue is to identify \textit{where} these dimensions are located in the body. For instance, even domain experts like tailors and anthropometrists can provide human body dimensions with limited precision.
\begin{figure}[H]
	\begin{center}
		\includegraphics[width=\linewidth]{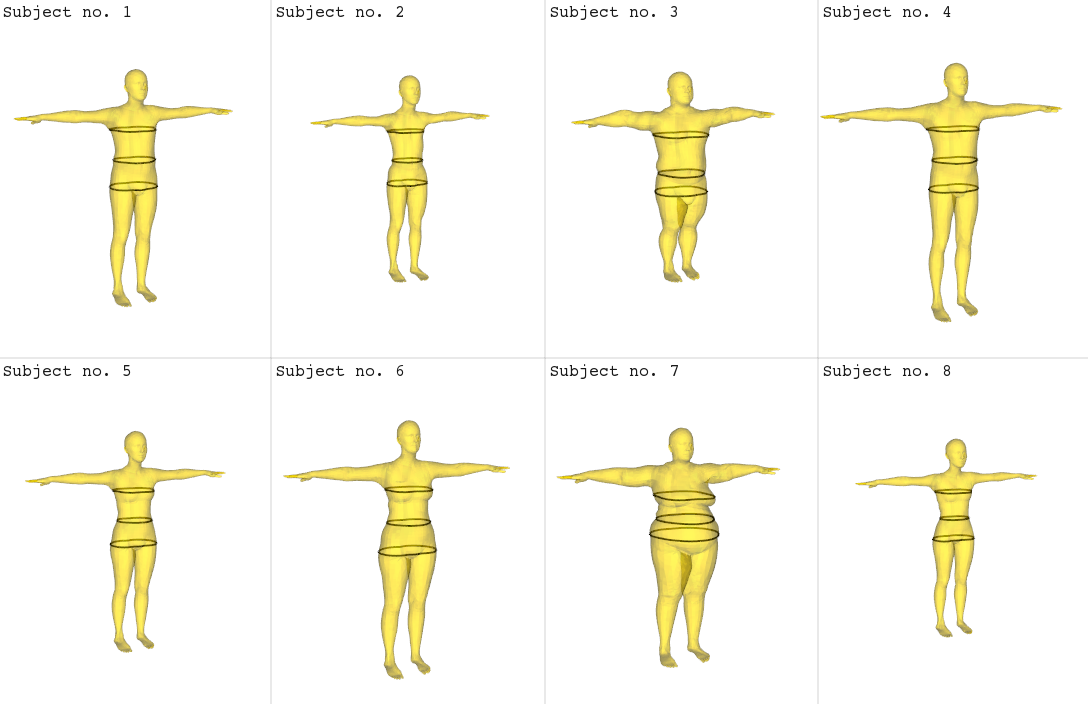}
	\end{center}
	\caption{Qualitative evaluation for male and female meshes.}
	\label{fig:qualitative_eval}
\end{figure}

Figure \ref{fig:qualitative_eval} shows the calculated dimensions on eight 
subjects, male subjects no. 1-4 and female subjects no. 5-8. The black paths 
represent chest, waist and pelvis circumference (from 
top to bottom) as defined by equations \ref{eq:cc}-\ref{eq:pc}, respectively. 
We slightly rotate the meshes and add transparency for visualization purposes. 
Note that our method is able to calculate automatically the human body 
dimensions for all 
subjects. The dimensions are indeed calculated at locations where domain 
experts regularly identify them.

\subsection{Learning with Calvis-Net}\label{subsec:learning}
This experiment demonstrates the plausibility of employing CALVIS to obtain 
data 
that can be used as a ground truth for machine learning (in general) and particularly as supervising signal for CNN.

\subsubsection{Experimental setup.} First, we assemble our Calvis-CMU Dataset. 
Using Blender we render 3400 grayscale images (1700 of each gender) of 
human body meshes generated with parameters from the CMU dataset. To increase 
the realism of these images we 
adopt an indoor lighting model with one light source and white background. This makes sense because it resembles the way 
how a person is standing in a room when she attends to the tailor for her 
body dimensions to be taken.
Subsequently, we annotate the CMU dataset with CALVIS. The 
automatic annotation process took approximately $1\,h\, 47\, min$ in an 
enhanced modern personal computer.

\begin{figure}[H]
	\begin{center}
		\includegraphics[width=\linewidth]{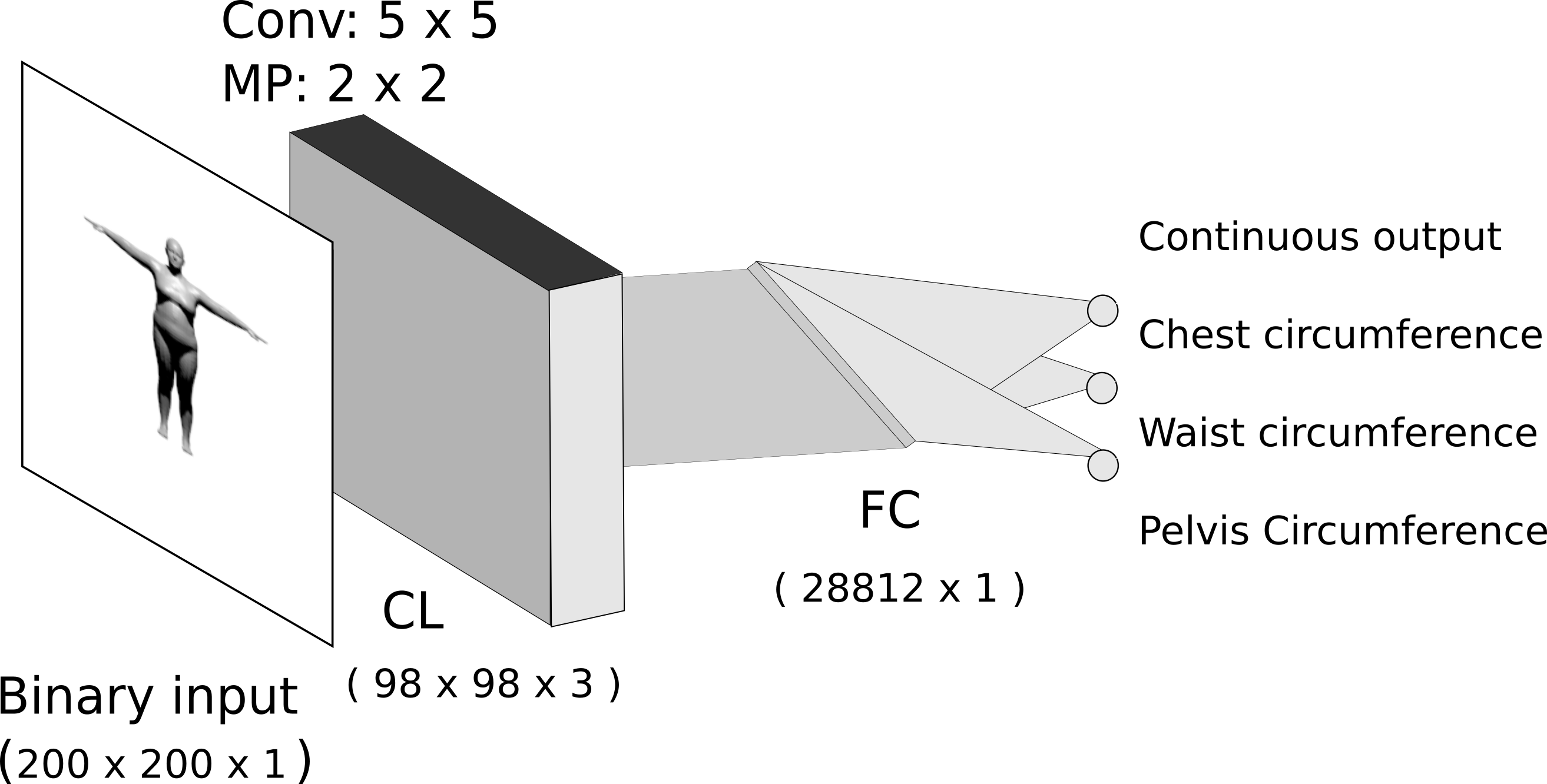}
	\end{center}
	\caption{Calvis-Net.}
	\label{fig:calvis_net}
\end{figure}

Figure \ref{fig:calvis_net} shows our prototypical CNN Calvis-Net implemented 
with pytorch \cite{paszke2017automatic}. The 
input is a synthetic grayscale image from a 3D human mesh of size $\mathtt{200 
x 200 x 
1}$. The input layer is followed by a convolutional layer $\mathtt{CL}_1$ with 
$\mathtt{3}$ 2D-kernels of 
size $\mathtt{5 x 5}$ and max pooling $\mathtt{MP}$ over a $\mathtt{2 x 2}$ 
volume. The network continues then with the rectified linear activation function
(ReLU). The last layer (fully connected $\mathtt{FC}$) regresses directly three 
body 
dimensions: chest, waist and pelvis circumference. We train the network for 20 
epochs and perform 3-fold 
cross-validation, achieving mean absolute validation error of 12 
mm.

%------------------------------------------------------------------------
\section{Conclusion}
In this paper we presented CALVIS, a method to calculate chest, waist and pelvis circumference from 3D human meshes. 
We demonstrate that our approach can be used to train a CNN by imputing synthetic images of humans and using the measurements calculated with CALVIS as supervision signal.
Furthermore, we contribute with a prototype CNN CALVIS-NET. Our experiments show that the CNN is able to learn how to perform this task, achieving competitive mean absolute validation error.
The code 
and data use 
in this paper will be made publicly available for researchers.
\bibliographystyle{splncs04}
\bibliography{egbib}

\end{document}